\documentclass[10pt,journal,compsoc]{IEEEtran}

\usepackage{cite}
\ifCLASSINFOpdf
  \usepackage[pdftex]{graphicx}
\else
  \usepackage[dvips]{graphicx}
\fi
\usepackage{amsmath,amssymb,amsfonts}
\usepackage{array}

\usepackage{hyperref}
\usepackage{fancyhdr}
\usepackage{booktabs} 
\usepackage{multirow,multicol}
\usepackage{color,xcolor}
\usepackage{graphicx}
\usepackage{epstopdf}
\usepackage{booktabs}
\usepackage{cite}
\usepackage[Symbol]{upgreek}
\usepackage{psfrag}
\usepackage{setspace}
\usepackage{acronym}
\usepackage{arydshln}

\usepackage{bm}
\usepackage{verbatim}

\usepackage{algorithm}
\usepackage{algorithmicx}
\usepackage{algpseudocode}

\ifCLASSOPTIONcompsoc
  \usepackage[caption=false,font=normalsize,labelfont=sf,textfont=sf]{subfig}
\else
  \usepackage[caption=false,font=footnotesize]{subfig}
\fi

\usepackage{stfloats}

\ifCLASSOPTIONcaptionsoff
  \usepackage[nomarkers]{endfloat}
 \let\MYoriglatexcaption\caption
 \renewcommand{\caption}[2][\relax]{\MYoriglatexcaption[#2]{#2}}
\fi

\usepackage{url}

\DeclareMathAlphabet{\mathsfbr}{OT1}{cmss}{m}{n}
\SetMathAlphabet{\mathsfbr}{bold}{OT1}{cmss}{bx}{n}
\DeclareRobustCommand{\msf}[1]{%
  \ifcat\noexpand#1\relax\msfgreek{#1}\else\mathsfbr{#1}\fi
}

\makeatletter
\newcommand{\msfgreek}[1]{\csname s\expandafter\@gobble\string#1\endcsname}
\makeatother

\DeclareFontEncoding{LGR}{}{} 
\DeclareSymbolFont{sfgreek}{LGR}{cmss}{m}{n}
\SetSymbolFont{sfgreek}{bold}{LGR}{cmss}{bx}{n}
\DeclareMathSymbol{\salpha}{\mathord}{sfgreek}{`a}
\DeclareMathSymbol{\sbeta}{\mathord}{sfgreek}{`b}
\DeclareMathSymbol{\sgamma}{\mathord}{sfgreek}{`g}
\DeclareMathSymbol{\sdelta}{\mathord}{sfgreek}{`d}
\DeclareMathSymbol{\sepsilon}{\mathord}{sfgreek}{`e}
\DeclareMathSymbol{\szeta}{\mathord}{sfgreek}{`z}
\DeclareMathSymbol{\seta}{\mathord}{sfgreek}{`h}
\DeclareMathSymbol{\stheta}{\mathord}{sfgreek}{`j}
\DeclareMathSymbol{\siota}{\mathord}{sfgreek}{`i}
\DeclareMathSymbol{\skappa}{\mathord}{sfgreek}{`k}
\DeclareMathSymbol{\slambda}{\mathord}{sfgreek}{`l}
\DeclareMathSymbol{\smu}{\mathord}{sfgreek}{`m}
\DeclareMathSymbol{\snu}{\mathord}{sfgreek}{`n}
\DeclareMathSymbol{\sxi}{\mathord}{sfgreek}{`x}
\DeclareMathSymbol{\somicron}{\mathord}{sfgreek}{`o}
\DeclareMathSymbol{\spi}{\mathord}{sfgreek}{`p}
\DeclareMathSymbol{\srho}{\mathord}{sfgreek}{`r}
\DeclareMathSymbol{\ssigma}{\mathord}{sfgreek}{`s}
\DeclareMathSymbol{\stau}{\mathord}{sfgreek}{`t}
\DeclareMathSymbol{\supsilon}{\mathord}{sfgreek}{`u}
\DeclareMathSymbol{\sphi}{\mathord}{sfgreek}{`f}
\DeclareMathSymbol{\schi}{\mathord}{sfgreek}{`q}
\DeclareMathSymbol{\spsi}{\mathord}{sfgreek}{`y}
\DeclareMathSymbol{\somega}{\mathord}{sfgreek}{`w}

\DeclareMathSymbol{\svarsigma}{\mathord}{sfgreek}{`c}

\DeclareMathSymbol{\sGamma}{\mathalpha}{sfgreek}{`G}
\DeclareMathSymbol{\sDelta}{\mathalpha}{sfgreek}{`D}
\DeclareMathSymbol{\sTheta}{\mathalpha}{sfgreek}{`J}
\DeclareMathSymbol{\sLambda}{\mathalpha}{sfgreek}{`L}
\DeclareMathSymbol{\sXi}{\mathalpha}{sfgreek}{`X}
\DeclareMathSymbol{\sPi}{\mathalpha}{sfgreek}{`P}
\DeclareMathSymbol{\sSigma}{\mathalpha}{sfgreek}{`S}
\DeclareMathSymbol{\sUpsilon}{\mathalpha}{sfgreek}{`U}
\DeclareMathSymbol{\sPhi}{\mathalpha}{sfgreek}{`F}
\DeclareMathSymbol{\sPsi}{\mathalpha}{sfgreek}{`Y}
\DeclareMathSymbol{\sOmega}{\mathalpha}{sfgreek}{`W}

\DeclareRobustCommand{\mcal}[1]{%
  \ifcat\noexpand#1\relax\mathnormal{#1}\else\cal{#1}\fi
}
\DeclareRobustCommand{\BM}[1]{%
  \ifcat\noexpand#1\relax\bm{\boldUppercaseItalicGreek{#1}}\else\bm{#1}\fi
}
\makeatletter
\newcommand{\boldUppercaseItalicGreek}[1]{\csname var\expandafter\@gobble\string#1\endcsname}
\makeatother
\newcommand{\rv}[1]{\MakeLowercase{\msf{#1}}}
\newcommand{\RV}[1]{\bm{\MakeLowercase{\msf{#1}}}}

\newcommand{\V}[1]{\bm{#1}}

\begin{document}
%
\title{IIns-VAE+: A Robust Transfer Learning Framework for Environmental Identification in Wireless Sensing}
%
%

\author{Yuxiao~Li, 
        Keke~Hu,
        Bobai~Zhao,
        Santiago~Mazuelas,
        and~Yuan~Shen,
        \IEEEcompsocitemizethanks{\IEEEcompsocthanksitem Yuxiao Li and Santiago Mazuelas are with the Basque Center for Applied Mathematics (BCAM), 48009 Bilbao, Spain (e-mail: yli@bcamath.org, smazuelas@bcamath.org).}
        \IEEEcompsocitemizethanks{\IEEEcompsocthanksitem Keke Hu is with the College of Semiconductor (College of Integrated Circuits), Hunan University, Changsha 410082, China (e-mail: hukeke@hnu.edu.cn)}
        \IEEEcompsocitemizethanks{\IEEEcompsocthanksitem Bobai Zhao is with the College of Computer Science, Beijing Information Science and Technology University, Beijing 100192, China (e-mail: zhaobobai@bistu.edu.cn)}
        \IEEEcompsocitemizethanks{\IEEEcompsocthanksitem Yuan Shen is with the Department of Electronic Engineering and the Beijing National Research Center for Information Science and Technology, Tsinghua University, Beijing 100084, China, and also with the Shanghai AI Laboratory, Shanghai 201112, China (e-mail: shenyuan\_ee@tsinghua.edu.cn).}}

\IEEEtitleabstractindextext{
\begin{abstract}

Environmental identification in wireless sensing is essential for 6G integrated sensing
and communication (ISAC) systems to achieve reliable situational awareness.
However, deep learning (DL) models for this task often fail to generalize under domain shift across diverse environments.
While the Inter-Instance Variational Auto-encoder (IIns-VAE) learns features of rich representation, its neural classifier remains vulnerable to these distribution changes.
In this paper, 
we propose IIns-VAE+, a hybrid model that combines the IIns-VAE framework with Minimax Risk Classifiers (MRC) to improve adaptability in transfer learning scenarios.
We use real-world datasets to evaluate our framework across three transfer learning scenarios, including general to specific room environments, high to low label resolutions, and mixed to specific environments.
The experimental results indicate that IIns-VAE+ significantly outperforms baselines, demonstrating its critical value in building adaptable and robust perceptive networks in future 6G systems.

\end{abstract}

}

\maketitle

\IEEEdisplaynontitleabstractindextext

\IEEEpeerreviewmaketitle

%
\IEEEpeerreviewmaketitle

\IEEEraisesectionheading{
\section{Introduction}
\label{sec:intro}}

\IEEEPARstart{W}{ireless} sensing has been widely recognized as a key enabler of sixth-generation (6G) wireless systems, driven by their critical role in emerging applications ranging from smart homes to industrial automation \cite{deLima.Belot.etal2020, Trevlakis.Boulogeorgos.etal2023, J17:WanGaoMao,J19:WanLiuYan,MenMeyBauWin:J19}.
By sharing spectrum, hardware, and signal processing, integrated sensing and communication (ISAC) can improve resource utilization, enhance coordination, and increase overall network efficiency, thus enabling novel, immersive and context-aware experiences \cite{PinTan.He.etal2021, Wu.Wang.etal2025}. Within such systems, environmental identification is essential to ensure reliable and precise sensing for various scenarios \cite{DecOrdFerHe:J18}. 
The ability of environmental identification enables ISAC systems to dynamically adapt to changing conditions, thereby strengthening the core functionalities, e.g., object detection, real-time decision-making, and high-accuracy localization \cite{WinSheDai:J18,WinDaiShe:J18,SheWymWin:J10,RicFre:J17}.
These techniques underpin a wide range of advanced wireless applications,
including autonomous driving \cite{J18:SimKloAsa}, crowd sensing \cite{FlaAnd:J16}, environmental monitoring \cite{DarAndChi:J07}, and smart cities \cite{J17:YaqHas,J22:JavSha}.
Consequently, robust and accurate environmental identification is becoming increasingly vital for next-generation wireless systems, helping to pave the way for a new era of connectivity in beyond-5G (B5G) and 6G networks.

\begin{figure}[tb]
\centering
\includegraphics[width=0.45\textwidth]{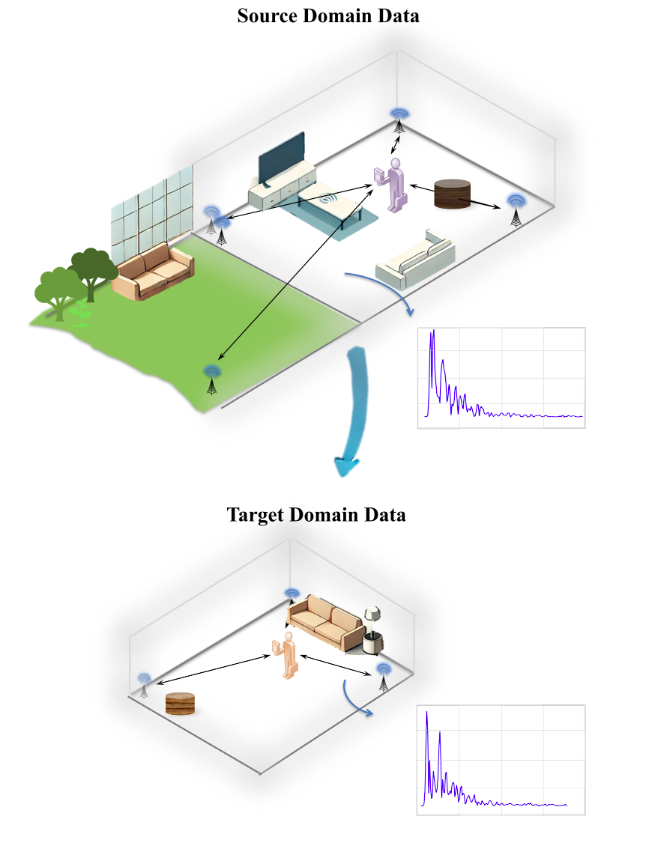}
\caption{Transfer learning in environmental identification. Target domain data can be collected from different environments compared to source domain data. 
}
\label{fig:intro}
\end{figure}

Environmental identification faces inherent challenges due to the variability in wireless scenarios.
Signal propagation is significantly affected by reflection, scattering, and attenuation caused by dynamic room layouts, heterogeneous material properties of obstacles, and varied sensor deployment configurations. These environmental variations directly manifest in the statistical properties of both the channel impulse response (CIR) and received waveform data, complicating the task of robust and reliable environmental identification in practical applications.
Existing environmental identification methods including traditional techniques such as Support Vector Machines (SVM) \cite{MarGifWymWin:J10,WymMarGifWin:J12}, Gaussian process regression (GPR) \cite{XiaWenMar:C13}, and more advanced deep learning (DL) models such as CNNs\cite{DecOrdFerHe:J18,LiMazShe:C22,LiMazShe:C22_2}, have demonstrated promising results in controlled scenarios.
However, these approaches are prone to overfitting to specific training conditions, which significantly limits their ability to generalize to unseen environments with different propagation characteristics \cite{MazConAllWin:J18}.
This limitation becomes particularly severe when the training of a model is confined to simplistic and static conditions, such as a single room configuration or a fixed set of obstacles, while its deployment demands robustness across a wide spectrum of dynamic and real-world environments.
Furthermore, many current models focus primarily on coarse-grained classification tasks, e.g., line-of-sight (LOS) versus non-line-of-sight (NLOS) discrimination \cite{MarGifWymWin:J10, WymMarGifWin:J12, XiaWenMar:C13}. 
This narrow scope limits their capability to extract more fine-grained environmental attributes, such as room geometry or material composition, which are essential for robust performance in complex applications requiring detailed situational awareness.


Transfer learning enables models to leverage knowledge from a source environment and adapt it to a target environment with minimal performance degradation. In the context of wireless sensing, particularly for ISAC systems, this paradigm allows models trained in a specific setting (e.g., a known indoor room) to generalize effectively to unseen environments (e.g., unfamiliar buildings or outdoor) without requiring extensive retraining \cite{NgyVanChu:J22}.
By reducing the reliance on large labeled datasets for each potential target domain, transfer learning substantially improves the scalability and practical viability of perceptive networks.
While common techniques (e.g., fine-tuning, adversarial domain adaptation, and meta-learning) have successfully addressed domain adaptation challenges in several applications such as signal detection \cite{VanLi:J22,LiuWeiNg:J21} and localization \cite{SunCheQi:C08,PanZhe:C08,LiuZhaNg:J17}, they often struggle with environmental identification, which is a key component of ISAC situational awareness. 
This difficulty stems from the substantial variability and complex distributional shifts in wireless environments, caused by the subtle alterations in signal patterns induced by structural or material disparities in the target environment \cite{AkrFerBel:J23, WanLinTia:J21}.

Meanwhile, the efficacy of several aforementioned methods is constrained by the underlying assumption of moderate distributional shifts, rendering them inadequate for the complex scenarios envisaged for 6G, including adaptation from dense urban canyons to deep indoor environments or across diverse office layouts. For instance, while transfer learning is employed to adapt localization models in \cite{WanLinTia:J21}, its reliance on feature alignment techniques remains effective only under minor domain shifts. Similarly, fine-tuning is investigated for cross-domain signal detection \cite{LiuWeiNg:J21}, yet performance deteriorates significantly in highly dynamic settings. Adversarial training \cite{AkrFerBel:J23}, though moderately successful in aligning feature distribution across source and target domains, fails to preserve fine-grained environmental characteristics, which are essential for detailed environmental identification in perceptive networks. More recently, DL models, like Inter-Instance Variational Auto-Encoder (IIns-VAE) \cite{LiMazShe:J23}, have demonstrated superior performance in environmental identification by effectively disentangling position-related and environment-related features. Nevertheless, its neural classifier exhibits limited generalization capability in unseen environments, constraining its practicality in transfer learning scenarios that are critical for scalable ISAC systems operating under inherent domain shifts.

In this paper, we propose IIns-VAE+, a hybrid model designed to advance transfer learning by integrating the feature extraction capability of IIns-VAE with the classification robustness of Minimax Risk Classifiers (MRC). The MRC is formulated to minimize the worst-case risk, enhancing resilience to domain shifts and ensuring adaptability to unseen environments  \cite{NEURIPS2023_cf4114c3, pmlr-v202-segovia-martin23a, pmlr-v162-alvarez22a}.
By incorporating both linear and Fourier mappings, MRC effectively captures complex environmental characteristics, thereby improving generalization across diverse practical environments \cite{pmlr-v216-bondugula23a}. This integration enables IIns-VAE+ to harness the complementary strengths of DL and risk-aware classification, ensuring reliable performance under varying environmental conditions and label resolutions.


The main contributions of this paper are summarized as follows:

\begin{itemize}
    \item We propose IIns-VAE+, a novel approach that integrates MRC into the IIns-VAE architecture.  This hybrid design synergistically combines the rich representational learning of DL and the worst-case robustness of MRC, significantly improving environmental identification performance in wireless sensing. 
    
    \item We introduce and formalize two flexible transfer learning strategies, layer-wise integration and bottleneck-wise integration. These strategies not only enable effective fine-tuning of the classifier for specific environments, but also provide scalable adaptation mechanisms under varying training conditions.
    
    \item We conduct a comprehensive evaluation of IIns-VAE+ in three real-world transfer learning scenarios and compare its performance with several baseline methods. Experimental results demonstrate that IIns-VAE+ achieves significant performance improvements, confirming its superior adaptability and generalization capabilities across diverse environments.
    
\end{itemize}

The remaining sections are organized as follows. Section~\ref{sec:related} presents the related work. Section~\ref{sec:problem} formulates the problem of transfer learning in wireless sensing, specifically for environmental identification.
Section~\ref{sec:method} presents the proposed IIns-VAE+ framework, detailing the integration of IIns-VAE with MRC and the two transfer learning strategies.
Section~\ref{sec:exp} provides a comprehensive experimental evaluation of the proposed framework using real-world datasets.
Finally, Section~\ref{sec:con} concludes the paper and discusses potential research directions.


\textit{Notations:} random variables (RVs) are displayed in sans serif, upright fonts and their realizations in serif, italic fonts; vectors are denoted by bold lowercase letters; a RV and its realization are denoted by $\rv{x}$ and $x$; a random vector and its realization are denoted by $\RV{x}$ and $\V{x}$; $\V{x}[j]$ denotes the $j$th component of the vector $\V{x}$; $p(\V{x}|y)$ denotes the conditional distribution of $\RV{x}$ given $\rv{y}=y$; $\mathcal{N}(\V{x};\boldsymbol{\mu}, \boldsymbol{\Sigma})$ denotes the PDF of a Gaussian RV $\RV{x}$ with mean $\boldsymbol{\mu}$ and covariance matrix $\boldsymbol{\Sigma}$;
$\mathbb{E}\{\cdot\}$ denotes the expectation of the argument, and $\mathbb{E}_{\RV{x}}\{\cdot\}$ denotes the expectation with respect to RV $\RV{x}$;
sets are denoted by calligraphic fonts, e.g., $\mathcal{D}$.


\section{Related Work}
\label{sec:related}

The rapid evolution of wireless sensing technologies, particularly in the context of 6G ISAC systems, has spurred significant research into robust environmental identification methods. These methods are essential to enable reliable situational awareness in dynamic and heterogeneous settings. In this section, we review two main strands of related work: (1) DL-based approaches that leverage channel state information (CSI) for sensing and localization, and (2) variational inference (VI) techniques that model uncertainty and facilitate robust inference in complex wireless environments. Together, these lines of research provide the foundation for our proposed IIns-VAE+ framework.

\subsection{Deep Learning-based Approach}
Deep learning has significantly advanced indoor sensing and localization by extracting complex patterns directly from raw  CSI, overcoming the limitations of traditional model-based methods. This part explores the mainstream deep learning architectures repurposed for this task.

\begin{table*}[tb]\centering
\caption{Summary of Deep Learning-Based Indoor Localization Algorithms}
\scriptsize
\label{tab:DL_sum}
\resizebox{\linewidth}{!}{
\begin{tabular}{p{2.2cm} p{3.8cm} p{3.5cm} p{1.8cm}}
\toprule
\textbf{Category} & \textbf{Core Idea} & \textbf{Representative Model} & \textbf{References} \\
\midrule
\textbf{1. Based on Convolutional Neural Network (CNN)} & Treats CSI as an image (2D grid data) and uses CNN to extract its \textbf{spatial features} (e.g., beam direction, multipath structure). & \textbf{CSILoc}: Uses 1D CNN to process multi-channel (amplitude, phase) CSI, followed by global average pooling. & \cite{Wang.Pasricha2022} \\
 & & \textbf{Duloc}: Employs a dual-channel CNN, processing CSI divided into two branches based on subcarrier stability. &  \cite{Song.Zhou.etal2022} \\
\addlinespace
\textbf{2. Based on Recurrent Neural Network (RNN)} & Uses RNNs (e.g., LSTM, GRU) to process \textbf{temporal features} in CSI, capturing how the channel varies over time. & \textbf{Zhang et al.}: Uses an LSTM network to model preprocessed CSI amplitude and phase information, extracting temporal features. & \cite{Zhang.Qu.etal2020} \\
\addlinespace
\textbf{3. Based on Attention Mechanism} & Uses self-attention mechanisms to fuse information from different antennas, subcarriers, and time dimensions in CSI, better capturing long-range dependencies. & \textbf{LoT}: Leverages the Vision Transformer (ViT) concept, converting CSI into an image format processed by a standard Transformer model. & \cite{Li.Meng.etal2023} \\
 & & \textbf{Swin-Loc}: Uses Swin Transformer layers to extract features from processed CSI data. & \cite{Xu.Zhu.etal2024} \\
 & & \textbf{ACPNet}: Combines two types of attention mechanisms (self-attention and channel attention) to process CSI from MIMO systems. & \cite{Wan.Chen.etal2024} \\
\addlinespace
\textbf{4. Hybrid Models} & Combines multiple network architectures to extract different dimensional features from CSI simultaneously (e.g., spatial + temporal). & \textbf{Hi-Loc}: Combines CNN and Bidirectional LSTM (Bi-LSTM), with a dual-attention mechanism to extract spatial and temporal information respectively. & \cite{Ruan.Chen.etal2022} \\
\bottomrule
\end{tabular}
}
\end{table*}

\textbf{Convolutional Neural Networks (CNNs)} represent a foundational approach for CSI-based localization. These algorithms treat the multi-dimensional CSI data, with its antenna and subcarrier dimensions, as a pseudo-image. This allows CNNs to effectively extract spatial features such as beam directions and multipath structures using their inherent convolution and pooling operations. Models like CSILoc \cite{Wang.Pasricha2022} and Duloc \cite{Song.Zhou.etal2022} demonstrate the effectiveness of CNNs in achieving robust localization by learning these spatial correlations. However, a significant limitation of CNN-based methods is their inherent weakness in modeling temporal dependencies. They typically process each CSI snapshot independently, ignoring the valuable sequential information that arises from user movement, which can lead to reduced accuracy in dynamic tracking scenarios.

\textbf{Recurrent Neural Networks (RNNs)}, particularly long short-term memory (LSTM) networks, have been employed to address the temporal dynamics of wireless signals,. As demonstrated in the work by Zhang et al. \cite{Zhang.Qu.etal2020}, RNNs excel at processing sequential CSI data, capturing temporal features like periodic patterns and trends in the channel's evolution over time. This makes them highly suitable for tracking moving devices and improving localization smoothness. Despite this advantage, RNNs, including LSTMs, often struggle with capturing long-range dependencies within sequences due to issues like vanishing gradients. Furthermore, their sequential processing nature limits computational parallelism, making training and inference slower compared to other architectures, especially with long data sequences.

The \textbf{Attention Mechanism}, and specifically the Transformer architecture, has emerged as a powerful successor, overcoming many limitations of previous models. Algorithms like LoT \cite{Li.Meng.etal2023}, Swin-Loc \cite{Xu.Zhu.etal2024}, and ACPNet \cite{Wan.Chen.etal2024} leverage self-attention to globally model relationships across all dimensions of the CSI data—antennas, subcarriers, and time—simultaneously. This allows for a more comprehensive feature fusion, leading to state-of-the-art localization accuracy. The key strength of attention-based models is their ability to handle long-range dependencies effectively and their high parallelism. The primary drawback, however, is their high computational complexity and substantial data requirements. They often need large amounts of data to generalize well and avoid overfitting, and their resource demands can challenge deployment on resource-constrained devices.

Finally, \textbf{Hybrid Models} seek to combine the strengths of various architectures to create more powerful solutions. A prominent example is Hi-Loc \cite{Ruan.Chen.etal2022}, which integrates CNNs for spatial feature extraction and Bidirectional LSTMs for temporal modeling, augmented with attention mechanisms. This fusion aims to build a more holistic understanding of the CSI data by capturing both its spatial and temporal characteristics. While these models often achieve superior performance by design, their main limitation is increased model complexity and size. This complexity leads to longer training times, higher computational costs, and greater challenges in practical deployment and optimization, posing a trade-off between performance and efficiency.

\subsection{Variational Inference for Sensing and Localization}

Variational Inference (VI) has established itself as a powerful and flexible framework for addressing the inherent uncertainties and complex data distributions encountered in wireless signal processing. By moving beyond simplistic point estimates to model rich posterior distributions, VI enables the development of more robust, reliable, and intelligent communication and sensing systems. Related work are summarized as following: 

\begin{table*}[tb]
\centering
\caption{Summary of Variational Inference for Wireless Sensing and Localization}
\label{tab:vi_sum}
\scriptsize
\resizebox{\linewidth}{!}{
\begin{tabular}{p{2.8cm} p{4.2cm} p{3.5cm}p{1.5cm}}
\toprule
\textbf{Category} & \textbf{Core Idea} & \textbf{Representative Work(s)} & \textbf{Reference(s)} \\
\midrule

\textbf{1. Foundational Theories} & & & \\
\cmidrule{1-4}
\quad \textit{1.1 Variational Inference} & 
Approximate complex posterior distributions by optimizing a tractable variational distribution. & 
Variational Inference Framework & 
\cite{BleKucMcA:J17} \\
\addlinespace

\quad \textit{1.2 Deep Generative Models} & 
Learn complex data distributions using deep neural networks for generation and representation learning. & 
Variational Autoencoder (VAE) & 
\cite{KinWel:C13} \\
\addlinespace

& & 
Generative Adversarial Network (GAN) & 
\cite{Goodfellow2014GenerativeAN} \\
\addlinespace

\quad \textit{1.3 Implicit Distributions} & 
Model complex distributions without explicit density functions through stochastic transformations. & 
Implicit Distribution Models & 
\cite{huszar2017variational} \\
\addlinespace
\midrule

\textbf{2. Localization under Uncertainty} & & & \\
\cmidrule{1-4}
\quad \textit{2.1 Environment as Latent Variable} & 
Treat environmental conditions as hidden variables affecting signal observations. & 
Machine learning for UWB ranging error mitigation & 
\cite{WymMarGifWin:J12} \\
\addlinespace

\quad \textit{2.2 Probabilistic Distance Estimation} & 
Estimate full distance posterior distribution rather than point estimates. & 
Soft Range Information (SRI) & 
\cite{MazConAllWin:J18, ConMazBar:J19} \\
\addlinespace
\midrule

\textbf{3. Disentangled Representation Learning} & & & \\
\cmidrule{1-4}
\quad \textit{3.1 Feature Disentanglement} & 
Separate different factors of variation (distance vs. environment) in latent space. & 
VI-based LVM for signal feature disentanglement & 
\cite{LiMazShe:J23} \\
\addlinespace

\quad \textit{3.2 Joint Inference} & 
Perform multiple tasks simultaneously from a single input signal. & 
IIns-VAE for concurrent distance estimation and environment identification & 
\cite{LiMazShe:J23} \\
\addlinespace
\midrule

\textbf{4. Signal Generation \& Adaptation} & & & \\
\cmidrule{1-4}
\quad \textit{4.1 Realistic Signal Synthesis} & 
Generate high-fidelity wireless signals with specific distance and environment characteristics. & 
IIns-GAN with adversarial training for signal generation & 
\cite{Goodfellow2014GenerativeAN, LiMazShe:J23} \\
\addlinespace

\quad \textit{4.2 Domain Adaptation} & 
Adapt signals across different environmental conditions while preserving distance information. & 
IIns-GAN for cross-domain signal translation & 
\cite{Goodfellow2014GenerativeAN, huszar2017variational} \\
\bottomrule
\end{tabular}
}
\end{table*}

\textbf{Foundational Theories and Methodologies:} The application of VI in wireless sensing is built upon several core theoretical foundations. The VI framework itself \cite{BleKucMcA:J17} provides the fundamental mechanism for approximating intractable posterior distributions through optimization techniques. Deep generative models, including Variational Autoencoders (VAEs) \cite{KinWel:C13} and Generative Adversarial Networks (GANs) \cite{Goodfellow2014GenerativeAN}, offer powerful architectures for learning complex data distributions, while implicit distribution models \cite{huszar2017variational} enable effective handling of distributions without explicit density functions. The primary strength of this theoretical foundation lies in its ability to integrate Bayesian inference with the representational power of deep learning, creating a versatile framework for probabilistic modeling. However, these methods introduce considerable complexity in model design and training, with VI being susceptible to approximation errors from restrictive variational families, and GANs facing persistent challenges in training stability and mode collapse.

\textbf{Localization Under Environmental Uncertainty:}
A critical application of VI in wireless systems addresses localization challenges in environmentally uncertain conditions. This approach reconceptualizes environmental impacts by treating them as latent variables within a probabilistic framework \cite{WymMarGifWin:J12}, advancing beyond traditional point estimates to full posterior distribution estimation through concepts like Soft Range Information \cite{MazConAllWin:J18, ConMazBar:J19}. The key advantage of this paradigm is its capacity to explicitly quantify uncertainty, thereby providing systems with inherent robustness against outliers and adverse environmental conditions. By modeling complete distance posteriors instead of single-point estimates, these methods support more reliable decision-making in downstream applications. Limitations include the computational overhead of distribution estimation and sensitivity to model misspecification, where inaccuracies in modeling environmental effects can propagate through the inference pipeline.

\textbf{Disentangled Representation Learning for Joint Inference:}
The advancement toward disentangled representation learning marks significant progress in wireless signal processing. Through hierarchical latent variable models \cite{LiMazShe:J23}, this methodology separates different factors of variation—specifically distinguishing distance-related features from environmental characteristics—within the latent space. This disentanglement facilitates simultaneous execution of multiple tasks from a single input signal, such as concurrent distance estimation and environment identification \cite{LiMazShe:J23}. The approach offers substantial benefits including enhanced model interpretability, improved generalization through invariant representations, and efficient multi-task learning. Nevertheless, significant challenges remain, particularly in designing architectures that ensure clean factor separation and addressing the amortization gap problem that can degrade inference performance for out-of-distribution samples.

\textbf{Signal Generation and Domain Adaptation:}
The generative capabilities of VI-based models provide innovative solutions to the critical challenge of data scarcity in wireless systems. By leveraging advanced architectures that integrate VAEs with GANs \cite{Goodfellow2014GenerativeAN, LiMazShe:J23}, these methods can synthesize high-fidelity wireless signals with specific characteristics and adapt signals across diverse environmental conditions \cite{Goodfellow2014GenerativeAN, huszar2017variational} while maintaining essential information integrity. The advantages are substantial, including dramatic reduction in real-world data collection costs, comprehensive testing capabilities for rare scenarios, and enhanced algorithm development through dataset augmentation. The fundamental challenge lies in ensuring the physical realism and fidelity of generated signals, particularly in preserving critical signal properties and preventing the amplification of spurious correlations from training data.

In summary, while DL models have demonstrated strong capabilities in extracting discriminative features from wireless signals, they often lack robustness under domain shift. Meanwhile, VI offers a principled framework for handling uncertainty and facilitating transferable representation learning, yet its integration with risk-aware classification remains underexplored. Our work bridges these two paradigms by combining the disentangled representation power of IIns-VAE with the distributionally robust classification of MRC, thereby addressing key limitations in existing approaches for environmental identification under domain shift.


\section{Problem Statement}
\label{sec:problem}

This section formally presents the transfer learning problem for environmental identification in wireless sensing, specifically addressing the challenges posed by domain shifts.
Three crucial types of transfer scenarios are introduced to illustrate the complexities of transfer learning. These scenarios vary in both environmental conditions and signal distributions.
Our approach leverages the IIns-VAE for feature extraction and the MRC for robust classification under domain shifts.
Finally, we introduce the evaluation metrics and baseline methods to rigorously measure the framework's effectiveness.

\begin{figure*}[!ht]
\centering
\includegraphics[width=0.7\textwidth]{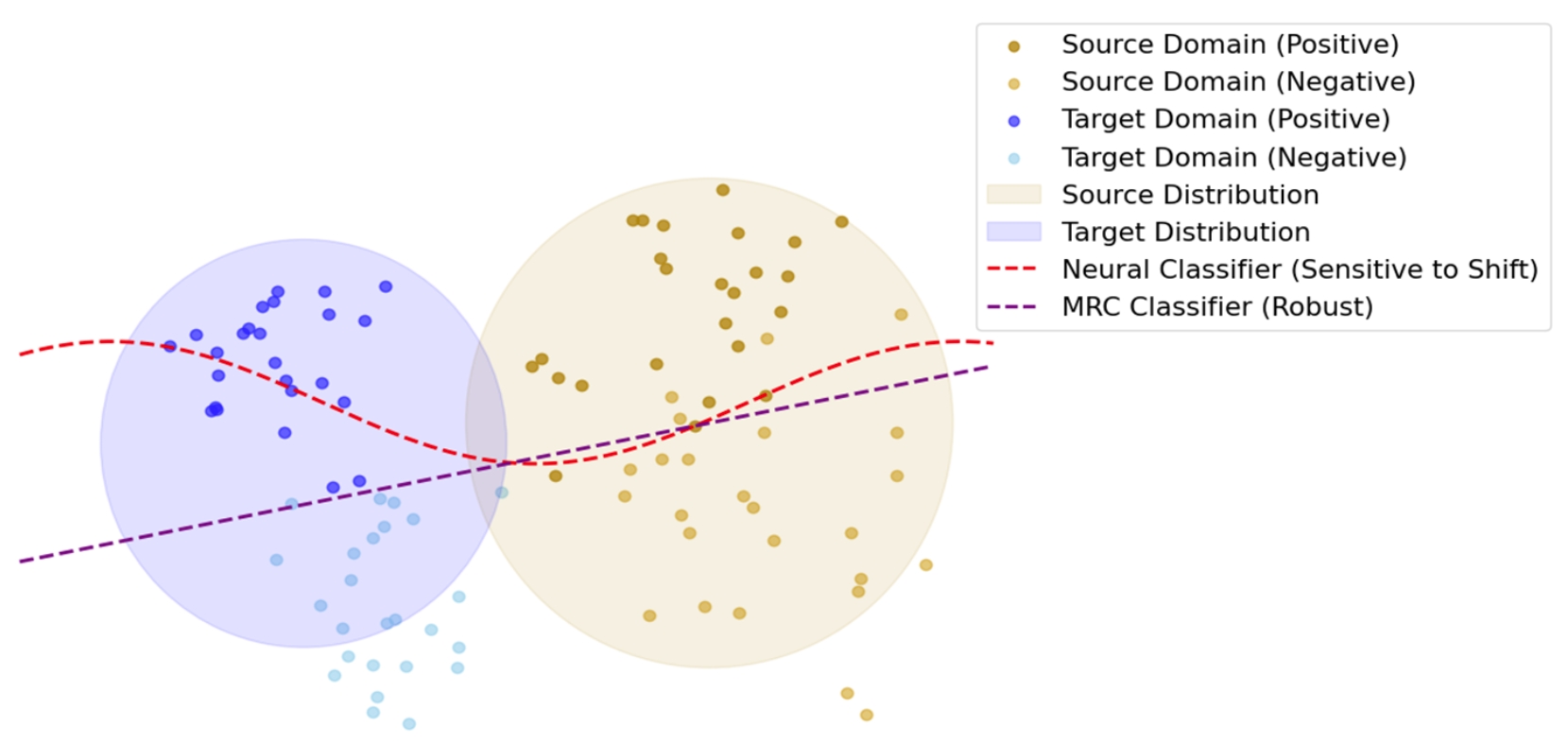}
\caption{Comparison of the neural classifier and the MRC classifier under domain shift. The MRC classifier provides resilience to domain shift and ensures more adaptability to new data domain.}
\label{fig:classifier_compare_1}
\end{figure*}

\subsection{Channel Impulse Response and Signal Representation}
 In ISAC systems, CSI serves not only for communication equalization but also as a perceptual cornerstone for interpreting the physical environment. Generally, the CSI of a multi-antenna, multi-carrier system is a complex matrix, $\mathbf{H}(f, t)$, containing multi-dimensional information in space, frequency, and time, providing rich representation to support advanced functions such as beamforming and resource allocation.

However, the development of robust perceptual capabilities must be grounded in an accurate understanding of the most fundamental properties of the channel, i.e., how multipath propagation in the time domain captures environmental characteristics. This is precisely the core value of the channel impulse response (CIR), as the core of its time dimension, which represents the most direct and fundamental source of information for interpreting environmental features. Therefore, for researching environmental identification, a fundamental ISAC task, an in-depth investigation into the characteristics of the CIR is particularly important and representative. The CIR, denoted as $h(t)$, has a physical model that clearly reveals the nature of the signal-environment interaction:

\begin{equation}
    h(t) = \sum_{i=1}^N a_i \delta(t-\tau_i)
\end{equation}
where $N$ represents the number of significant signal paths, $a_i$ is the amplitude of the $i$-th path, $\tau_i$ is the corresponding delay, and $\delta(\cdot)$ is the Dirac delta function. Each path is influenced by environmental properties such as geometry and material composition, rendering the CIR a high-dimensional descriptor of the spatial and material structure of the propagation environment.

Let $s(t)$ denote the transmitted signal, and assume the received signal $\V{x}(t)$ is the convolution of $s(t)$ with the CIR $h(t)$, combined with noise $n(t)$:
\begin{equation}
    \V{x}(t) = s(t) * h(t) + n(t)
\end{equation}
where $*$ denotes convolution. The received signal $\V{x}(t)$ thus carries information about both the transmitted signal and the environment through $h(t)$. By analyzing $\V{x}(t)$, we aim to infer environmental characteristics, including factors like LOS or NLOS conditions, room geometry, and the presence of obstacles.

For environmental identification, relevant features are extracted from the CIR and the received signal $\V{x}(t)$. The extracted features can be classified into position-related features and environment-related features. Position-related features, such as path delays $\tau_i$, are associated with distance and angle information while environment-related features, such as path amplitudes $a_i$ and signal scattering, are influenced by obstacles and material properties.

The CIR-based feature extraction forms the basis of our IIns-VAE model, where these features are mapped into latent variables for classification tasks. The robustness of this classification is further enhanced by the MRC, which mitigates the impact of domain shifts by minimizing the worst-case classification risk seen in Figure \ref{fig:classifier_compare_1}.

\subsection{Problem Definition: Transfer Learning for Environmental Identification}

In wireless sensing, environmental identification involves classifying the environmental scenario based on received signal measurements. Formally, let $k\in\mathcal{K} = \{0, 1, ..., K-1\}$ denote the environmental label of a scenario where a signal is measured, and $p \in [0, p_{max}]$ represents the positional parameter (e.g., distance, angle, etc.). The environmental label set $\mathcal{K}$ can vary in definition, depending on the classification requirements:
\begin{itemize}
    \item For binary conditions, such as LOS versus NLOS conditions, the environmental label is $k\in\mathcal{K}_{LOS} = \{0, 1\}$, where $k=0$ denotes LOS and $k=1$ denotes NLOS.
    \item In elaborated scenarios, $\mathcal{K}$ may represent multiple attributes, such as obstacle types $\mathcal{K}_{obs}$ or room geometries $\mathcal{K}_{room}$, where each element of $k$ captures a particular environmental characteristic. For instance, $k\in\mathcal{K}_{obs} \times \mathcal{K}_{room}$ could denote both the presence of obstacles and specific room characteristics.
\end{itemize}
The goal of environmental identification is to accurately classify the label $k$ based on signal data, distinguishing between conditions that impact signal propagation.

A major challenge in environmental identification is the domain shift between the training environment (source domain) and the operating environment (target domain). In real-world applications, models are often trained on data collected in one environment, where the distribution of signal measurements $P_S(\V{x}, k)$ is specific to that setting. However, when the model is deployed in a new, unseen environment (target domain), the distribution $P_T(\V{x}, k)$ of signal data differs due to variations in room geometry, obstacles, or material properties. This domain shift leads to performance degradation because the model struggles to generalize from the source domain to the target domain.

Transfer learning addresses the problem of domain shift by allowing a model trained on one environment (source domain) to adapt to a new environment (target domain) while minimizing performance degradation.
In a transfer learning framework, we define the source domain and target domain as follows:
\begin{itemize}
    \item \textbf{Source Domain} 
    $\mathcal{D}_S = \{(\V{x}_S^i, k_S^i)\}_{i=1}^{N_S}$: The source domain consists of $N_S$ labeled samples $(\V{x}_S^i, k_S^i)$ drawn from the joint distribution $P_S(\V{x}, k)$, where $\V{x}_S^i$ represents signal measurement and $k_S^i$ represents the corresponding environmental label. This domain provides sufficient labeled data for robust training
    \item \textbf{Target Domain} $\mathcal{D}_T = \{(\V{x}_T, k_T)\}$: The target domain consists of $N_T$ labeled samples $(\V{x}_T^j, k_T^j)$ from the distribution $P_T(\V{x}, k)$. Here, $N_T$ is typically much smaller than $N_S$, reflecting the practical constraint of limited labeled data in new environments. The domain shift between $P_S(\V{x}, k)$ and $P_T(\V{x}, k)$ introduces challenges in generalizing across domains.
\end{itemize}

The objective is to develop a model that can effectively use the abundant labeled data from the source domain distribution $P_S$ to achieve accurate environmental classification in $P_T$, despite the limited labeled samples in the target domain. This is essential for ensuring that environmental identification models can adapt to new settings with minimal additional data.

To comprehensively address transfer learning, we consider three types of transfer scenarios that highlight distinct challenges in environmental identifications:
\begin{enumerate}
    \item \textbf{Room Variability}: The model trained in one indoor environment with specific room layout and materials (source domain) is transferred to a new indoor environment with different geometry or materials (target domain). This scenario examines the model’s ability to generalize across spatially varying indoor settings.
    \item \textbf{Resolution Change}: The model trained on data sampled at a certain resolution (source domain) must adapt to data with a different sampling resolution (target domain). This scenario assesses the model’s robustness to scale variations and potential loss of information at different resolutions.
    \item \textbf{Mixed-to-Specific Environments}: The model is trained on a combined dataset covering various environments, such as both indoor and outdoor settings (source domain), and then applied to a specific, confined environment, like a single indoor setting (target domain). This setting tests the model’s adaptability from diverse, mixed-source training data to a constrained target scenario.
\end{enumerate}

The key problem, then, is to design a model that can perform accurate environmental identification across these scenarios by effectively adapting from $\mathcal{D}_S$ to $\mathcal{D}_T$ while preserving high classification accuracy despite distributional shifts.


\section{The IIns-VAE+ Framework}
\label{sec:method}


This section introduces the IIns-VAE+ framework, developed to address the limitations of traditional neural classifiers in transfer learning for environmental identification. By combining the IIns-VAE with MRC, the framework achieves robust classification under domain shifts.
We present the general methodology, integration strategies, transfer learning process, and a brief discussion on the framework's adaptability.

\subsection{Solution Components: IIns-VAE and MRC}

To takle the challenges of transfer learning for environmental identification, we employ two primary components: IIns-VAE\cite{LiMazShe:J23} for latent feature extraction and the MRC\cite{MazPer:C19} for risk-aware classification.

The IIns-VAE model is structured to separate the latent representations of position-related and environment-related features from wireless signals. Let $\V{x}$ represent the observed signal measurement, which contains both positional and environmental information. IIns-VAE seeks to model the latent variables $\V{z}_{\text{p}}$ (position-related) and $\V{z}_{\text{e}}$ (environment-related) by maximizing the evidence lower bound (ELBO) on the data log-likelihood for the instance-labels pair $(\V{x}, p, k)$:
\begin{equation}  \label{eq:bound_pro}
    \begin{aligned}
\mathcal{L}_{\text{ELBO}}(\V{x},d,k;\boldsymbol{\phi},\boldsymbol{\theta},\boldsymbol{\varphi})
    &= \mathbb{E}_{q(\V{z}_{\text{p}},\V{z}_{\text{e}}|\V{x};\boldsymbol{\phi})}\big\{\log p(\V{x}|\V{z}_{\text{p}},\V{z}_{\text{e}};\boldsymbol{\theta})\big\} 
    \\
    &\quad 
    - \operatorname{D}_{\text{KL}}\big(q(\V{z}_{\text{p}},\V{z}_{\text{e}}|\V{x};\boldsymbol{\phi})\big|\big|p(\V{z}_{\text{p}},\V{z}_{\text{e}})\big)  \\
    &\quad + \mathbb{E}_{q(\V{z}_{\text{p}}|\V{x};\boldsymbol{\phi})}\big\{\log p(p|\V{z}_{\text{p}};\boldsymbol{\varphi})\big\} \\
    &\quad + \mathbb{E}_{q(\V{z}_{\text{e}}|\V{x};\boldsymbol{\phi})}\big\{\log p(k|\V{z}_{\text{e}};\boldsymbol{\varphi})\big\}  \\
    &\leq \log\,p(\V{x}, d, k) \,.
    \end{aligned}
\end{equation}
where $q(\V{z}_{\text{d}},\V{z}_{\text{e}}|\V{x},d,k)$, $q(\V{z}_{\text{p}})$ and $q(\V{z}_{\text{e}}|\V{x})$ represent the approximate posterior distributions, $\operatorname{D}_{\text{KL}}$ denotes the Kullback-Leibler divergence, $\boldsymbol{\phi}$, $\boldsymbol{\theta}$, $\boldsymbol{\varphi}$ denote the distribution parameters to be inferred. Through this formulation, IIns-VAE captures structured latent representations that decouple environmental semantics $\V{z}_{\text{e}}$ from positional information $\V{z}_{\text{p}}$.

In traditional IIns-VAE, a neural classifier is applied to environmental features $\V{z}_{\text{e}}$ and generates the classification result. When the applied domains change, such neural classifiers struggle with generalization across domains. This lack of robustness in the neural classifier motivates our integration of a more resilient classification approach.

The MRC addresses the limitations of the neural classifier by minimizing worst-case classification risk. MRC seeks a classifer $h$ that optimizes the following optimization objective \cite{JMLR:v24:22-0339}: 
\begin{equation}
    h^* \in \arg \min_{h\in\mathcal{H}} \max_{p\in\mathcal{U}} \mathbb{E}_{p}\Big[l\big(h(\mathbf{x}), y\big)\Big]
\end{equation}
\noindent where $\mathcal{H}$ is the set of all possible classifiers, $\mathcal{U}$ is the set of all possible distributions over the data, and $l$ is a specified loss function.
By focusing on the worst-case distribution $p \in \mathcal{U}$, MRC ensures that the classifier performs robustly under varying distributions, making it more resilient to domain shift encountered in transfer learning.

The uncertainty set $\mathcal{U}$ is defined as a set of distributions $p \in \Delta(\mathcal{X}\times\mathcal{Y})$ that satisfy constraints on the expectations of a feature mapping $\Phi: \mathcal{X}\times\mathcal{Y}\to \mathbb{R}^{m}$. Specifically, the set is defined as:
\begin{equation}
    \mathcal{U} = \{p\in\Delta(\mathcal{X}\times\mathcal{Y}): |\mathbb{E}_{p}[\Phi(\V{x}, y)]-\tau|\leq\lambda\}
\end{equation}
where $\tau$ is the empirical mean vector of $\Phi$, and $\lambda$ is a confidence vector that controls the size of the uncertainty set based on the estimation errors in $\tau$. The feature mapping $\Phi$ is crucial for transforming data into a space where MRC can minimize the worst-case classification risk effectively.

MRC can employ different types of feature mappings $\Phi$. For example, linear mappings directly map features and are effective for scenarios with linearly separable features. Fourier feature mappings enable MRC to approximate non-linear decision boundaries in high-dimensional spaces by mapping inputs to a kernel-based freature space. This flexibility allows MRC to adapt to diverse environmental conditions, providing robust classification across different transfer settings\cite{pmlr-v162-alvarez22a}.

\begin{figure}[!bt]
\centering
\includegraphics[width=0.3\textwidth]{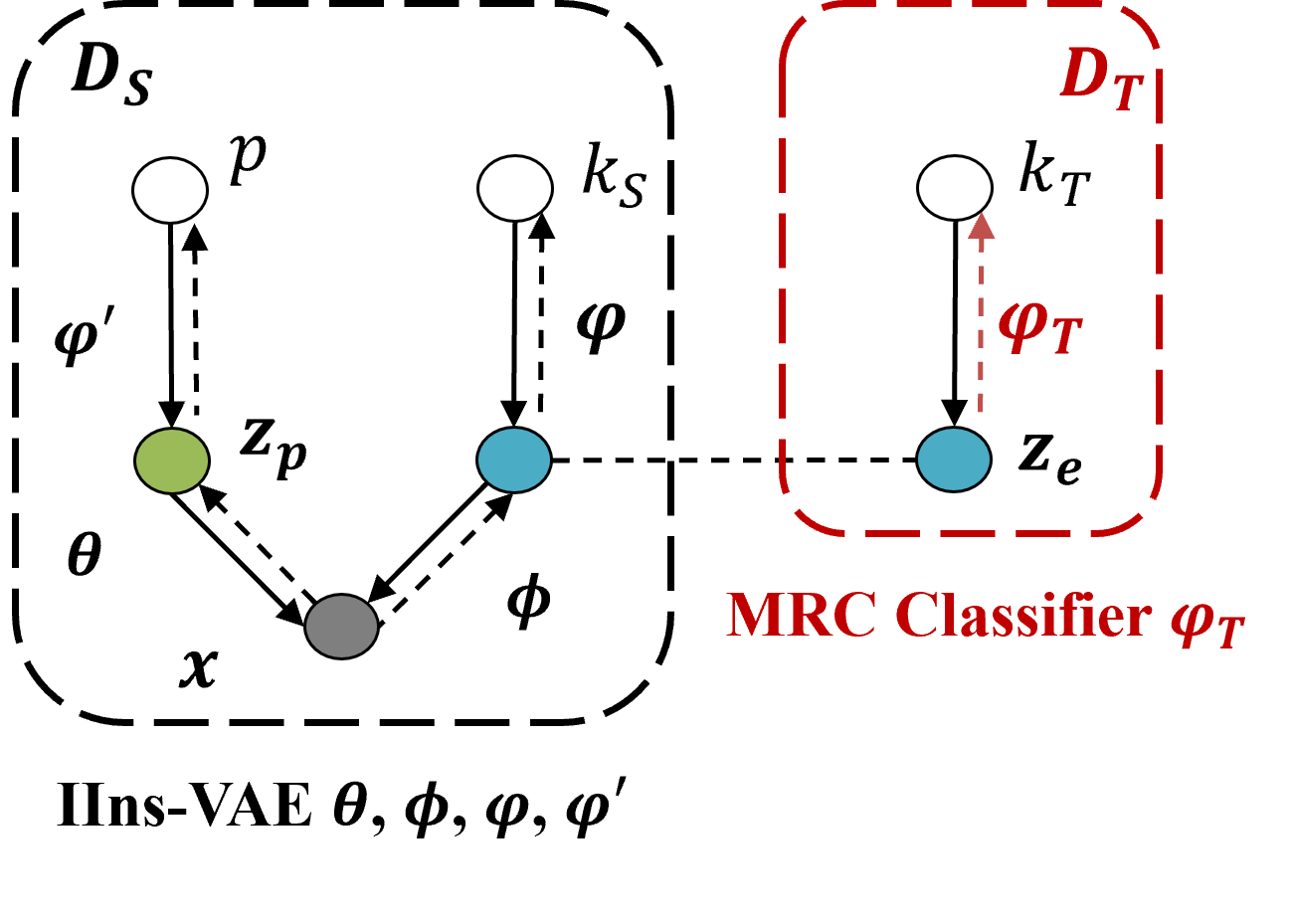}
\caption{Graphical model for IIns-VAE+ with the environment latent variables for the source and target domains.}
\label{fig:classifier_compare}
\end{figure}

The integration of MRC with IIns-VAE has the potential to address a fundamental limitation of traditional classifiers in transfer learning. While IIns-VAE excels at creating structured latent spaces through $\V{z}_{\text{p}}$ and $\V{z}_{\text{e}}$, its neural classifier's sensitivity to domain-specific variations limits its applicability in real-world settings with domain shifts, shown in Figure \ref{fig:classifier_compare}. In the following, we introduce our proposed IIns-VAE+.
By replacing the neural classifier with MRC, we combine the feature extraction power of IIns-VAE with the robustness of MRC for worst-case risk minimaiztion, enabling the model to generalize more effectively across domains.


\subsection{General Methodology: Combining IIns-VAE and MRC for Transfer Learning}

The IIns-VAE framework aims to enhance transfer learning performance in environmental identification by leveraging IIns-VAE's latent feature extraction and MRC's risk-aware classification.

In this combined model, IIns-VAE processes the received signal $\V{x}$ and decouples it into two latent variables, $\V{z}_{\text{p}}$ (position-related) and $\V{z}_{\text{e}}$ (environment-related). The distribution of $\V{z}_{\text{e}}$ encodes environmental features that serve as inputs for classification. This structure allows IIns-VAE to capture complex environmental patterns, which are essential for reliable environmental identification.

Instead of a traditional neural classifier, we apply MRC to $\V{z}_{\text{e}}$ for environmental classification. MRC minimizes the worst-case classification risk, ensuring that the classifier remains robust under distributional shifts. This approach enhances the model's generalization ability when applied to new domains.

The integration of IIns-VAE and MRC is designed to overcome the limitations of neural classifiers, providing a solution that remains effective across diverse environmental conditions with minimal retraining.

\subsection{Layer-wise and Bottleneck-wise Integration Strategies}

\begin{figure*}[htbp]
\centering
\includegraphics[width=0.95\textwidth]{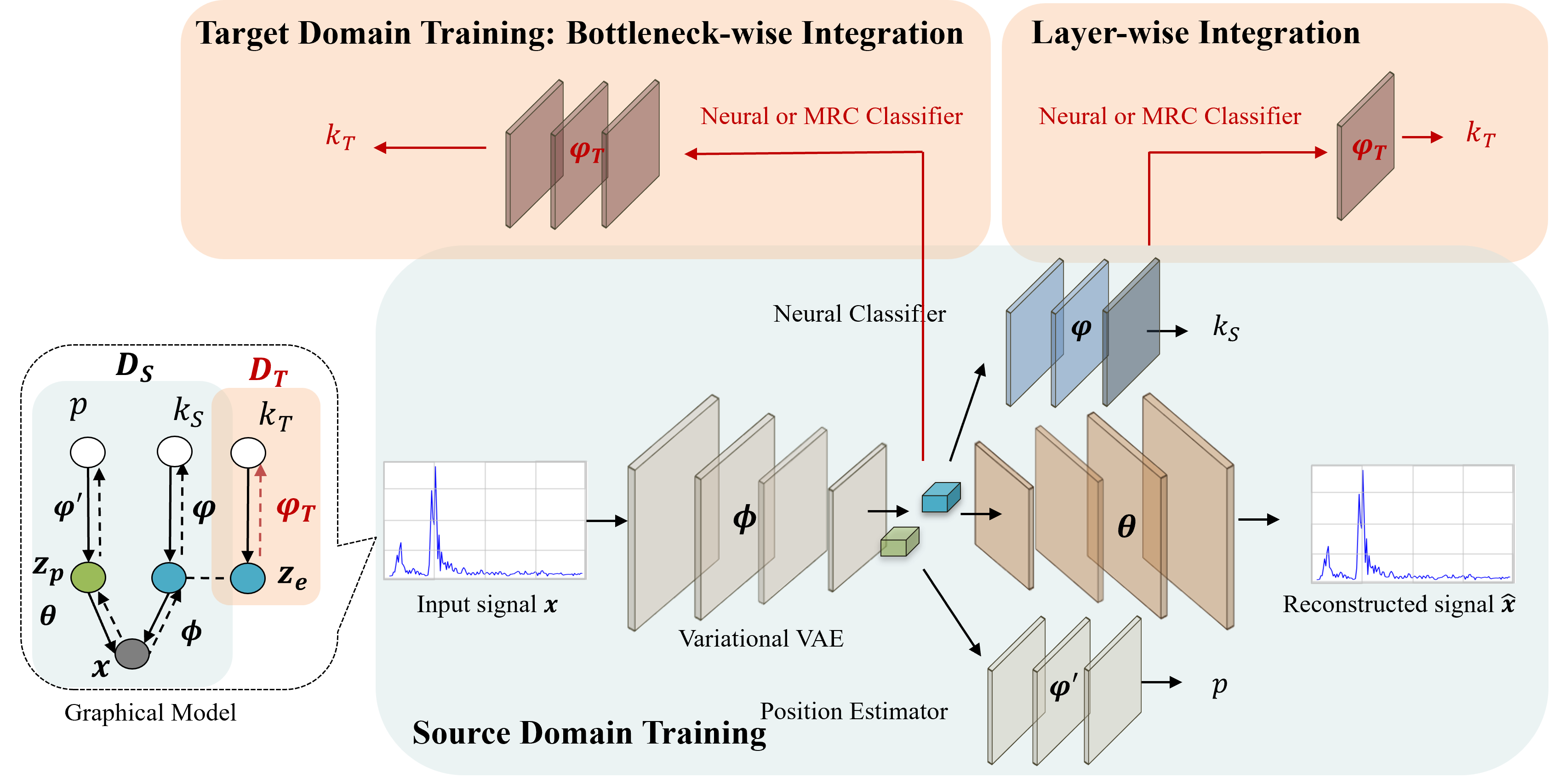}
\caption{Network structures of layer-wise and bottleneck-wise integration strategies.}
\label{fig:integrations}
\end{figure*}

To achieve adaptable transfer learning, the IIns-VAE+ framework incorporates two primary integration strategies. They are designed to fine-tune either the model's final layer or the bottleneck layer for the target domain.

\paragraph{Layer-wise Integration} In this scheme, the IIns-VAE model is trained on the source domain, after which all layers except the final layer are frozen during adaptation to the target domain. The final layer can then be adapted in one of the two ways:
\begin{enumerate}
    \item \textbf{Neural Network Layer}: The original neural classifier layer from IIns-VAE is fine-tuned to adjust the decision boundary for the target environment.
    \item \textbf{MRC Classifier}: Alternatively, the neural classifier can be replaced with the MRC classifier, which minimizes empirical risk under variable conditions. The MRC layer is then fine-tuned based on the last-layer features of IIns-VAE, providing robust, risk-aware classification in the target domain.
\end{enumerate}
This layer-wise integration approach is computationally efficient, as it updates only the final decision layer, making it well-suited for scenarios where the source and target domains are relatively similar.

\paragraph{Bottleneck-wise Integration} In this approach, the IIns-VAE model is initially trained on the source domain, and then the entire classifier, from the bottleneck to the output layer, is fine-tuned on the target domain. This enables a deeper adaptation by adjusting more parameters to fit the target domain's specific characteristics. The botleneck-wise integration strategy also supports two classifier options:
\begin{enumerate}
    \item \textbf{Neural Network Classifier}: The full original neural network classifier from IIns-VAE is fine-tuned, allowing for gradual adjustment through all classifier layers.
    \item \textbf{MRC Classifier}: The entire classifier is replaced with the MRC, enabling comprehensive risk-aware adaptation across layers. This approach is particularly beneficial for target environments that differ significantly from the source, as it allows deeper modifications to capture complex domain shifts.
\end{enumerate}

This dual-integration structure allows IIns-VAE+ to flexibly adapt to both minor and substantial domain shifts, enhancing the framework's overall adapatability for environmental identification across various wireless sensing scenarios.

\subsection{Transfer Learning Process}

The transfer learning process in the IIns-VAE+ framework is divided into two stages:
\begin{enumerate}
    \item \textbf{Source Domain Training}: \begin{itemize}
        \item The IIns-VAE model is trained on labeled source domain data $\mathcal{D}_{S}$ to learn a structured latent representation.
        \item The encoder processes $\V{x}_{S}$ to generate $\V{z}_{\text{p}}$ and $\V{z}_{\text{e}}$, where $\V{z}_{\text{e}}$ serves as the input to the neural classifier in IIns-VAE. The model is initially trained to minimize the reconstruction error and classification risk via \ref{eq:bound_pro} within the source domain distribution $P_S(\V{z}_{\text{p}})$.
    \end{itemize}
    \item \textbf{Target Domain Adaptation}:
    \begin{itemize}
        \item The IIns-VAE model is frozen, and is used to extract latent variables from target domain data $\mathcal{D}_{T}$.
        \item Depending on the complexity of domain shifts, either the layer-wise or bottleneck-wise integration strategy is chosen:
        \begin{itemize}
            \item In layer-wise integration, the MRC classifier is fine-tuned with IIns-VAE's last-layer feature on the target domain data.
            \item In bottleneck-wise integration, the MRC classifier is fine-tuned with IIns-VAE's bottleneck feature on the target domain data.
        \end{itemize}
    \end{itemize}
\end{enumerate}

These steps form the transfer learning algorithm, ensuring that the model adapts effectively to new environments with minimal computational cost.

\subsection{Discussion: Hybrid Approach for Transfer Learning}

The proposed IIns-VAE+ framework presents a flexible, hybrid approach with promising implications for transfer learning in wireless sensing. By combining deep generative modeling (IIns-VAE) with risk-aware classification (MRC), this framework tackles domain shifts inherent in environmental identification tasks. {\color{brown}This} makes it adaptable to dynamic and uncertain environments--a core challenge in real-world wireless sensing applications.

The layer-wise and bottleneck-wise integration serve as adaptable solutions, selectable based on the similarity between source and target environments. Layer-wise integration fine-tunes only the MRC classifier on the IIns-VAE's final-layer features. This preserves largely the encoder's learned information from the source domain, and allows efficient adaptation with minimal computational cost. This approach is ideal when environments share structural similarities. In such cases, it maintains stability while allowing modest adjustments to specific target features.

For feature mappings in MRC, we choose Fourier mappings for all cases.
For more complex shifts, bottleneck-wise integration enables a deeper adaptation by fine-tuning the MRC classifier on the bottlenecck features from IIns-VAE. Such integration strategy allows the model to capture nuanced differences in the target domain. Although computationally heavier, this strategy proves beneficial in environments with significant variabiliy, ensuring the model adapts robustly to challenging target conditions.

This hybrid approach leverages IIns-VAE's powerful latent representation learning, combined with MRC's robust decision-making under uncertainty. Such a combination allows the framework to generalize effectively across diverse environmental seetings. The flexibility of IIns-VAE+ extends further: It can incorporate other generative models for feature extraction or alternative classifiers for decision making. Such flexibility reveals its adaptability and potential as a versatile tool in transfer learning for more advanced wireless sensing problems.

\section{Experiments}
\label{sec:exp}



The experiments are designed to evaluate the proposed hybrid framework in practical transfer learning scenarios.
We compare the performance of IIns-VAE and its combination with MRC (using both linear and Fourier transformations) under two integration strategies across three distinct learning scenarios.
Accuracy, precision, recall and F1 scores are reported for each method and scenario.

\begin{figure}[tb]
    \centering
    \subfloat[PCA of Waveform Features by Room for Database 1]{%
        \includegraphics[width=0.45\textwidth]{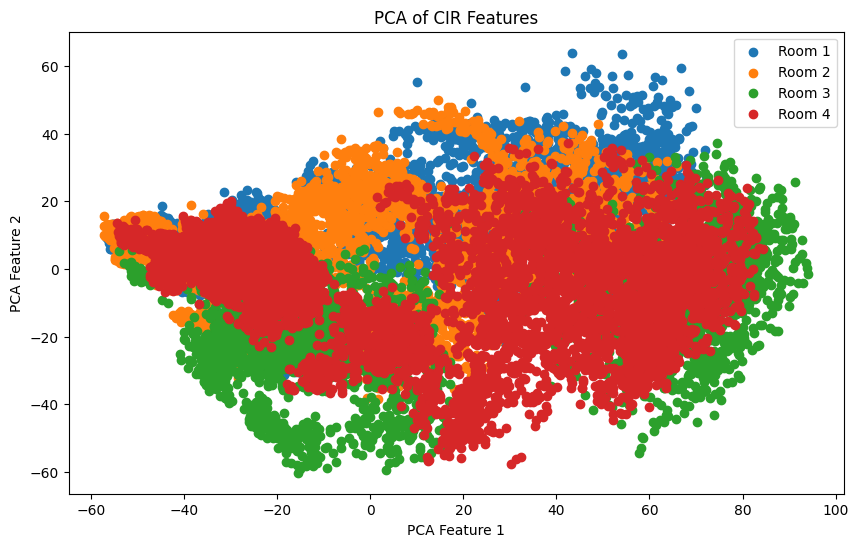}%
        \label{fig:pca_room_db1}
    }
    \hfill
    \subfloat[PCA of CIR Features by Room for Database 2]{%
        \includegraphics[width=0.45\textwidth]{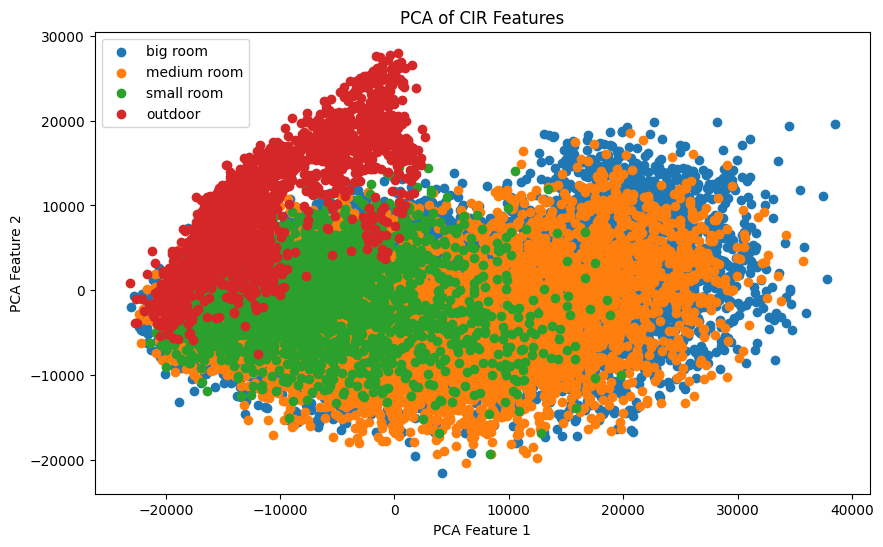}%
        \label{fig:pca_room_db2}
    }
    \caption{Comparison of PCA Feature Clustering Across Databases.}
    \label{fig:pca_comparison}
\end{figure}

\subsection{Datasets}

We explore the transfer learning capabilities of our proposed models using two comprehensive real-world datasets, introduced as follows:
\begin{itemize}
    \item \textbf{Database 1} is from \cite{Bregar2018ImprovingIL}, created using SNPN-UWB board with DecaWave DWM1000 UWB pulse radio module. Each sample includes signal measurements, a real distance, and an environmental label. The signal measurements per sample include a waveform of length $152$ and an estimated distance from the device. This database consists of two subsets:
    \begin{itemize}
        \item \textit{Database 1-1}: Includes $9,000$ samples recorded in adjacent office rooms and a hallway with LOS and NLOS environmental labels.
        \item \textit{Database 1-2}: Include $25,100$ samples recorded in a different office environment with multiple rooms, also labeled with LOS and NLOS conditions. 
    \end{itemize}
    \item \textbf{Database 2} is from \cite{zenodo}, created using DecaWave EVB1000 devices. Each sample includes signal measurements, a real distance, and an environmental label array. The signal measurements per sample include a waveform of length $157$ and an estimated distance from the device. The environmental label array includes two labels for room setting and blocking materials, respectively. This database consists of $49,233$ samples with signal measurements across various room settings (e.g., large office, small office, outdoor) and obstacle materials (e.g., metal, glass, wood, plastic).
\end{itemize}

These datasets represent diverse wireless channel environments, including indoor and outdoor spaces, and capture a wide range of channel impulse response (CIR) characteristics.
We further perform data analysis to investigate the data characteristics and design transfer learning scenarios. Specifically, we perform Principal Component Analysis (PCA) on the waveform features of two datasets. This analysis aims to visualize how well data clusters by environmental labels (i.e., room types and sizes) across different databases, as shown in Figure \ref{fig:pca_comparison}. Figure \ref{fig:pca_room_db1} presents the PCA feature clustering for Database 1, showing room-based clustering, while Figure \ref{fig:pca_room_db2} illustrates Database 2, with environmental types labeled by room size and outdoor settings. Despite some degree of clustering, there are significant overlaps between features from different rooms and environments. This overlap suggests that certain environmental features generalize across diverse settings, supporting the potential for transfer learning.

By linking these clusters to specific environmental types, we gain insight into how environmental factors influence signal characteristics. For example, larger room types may show distinct signal patterns compared to outdoor environments, indicating that source and target domain assignments could leverage these environmental distinctions. This label-based comparison across datasets highlights the relevance of transfer learning. It suggests that certain learned features can be adaptable across similar enviornmental types in unseen domains.

The insights help the design of transfer learning scenarios. Specifically, we aim to design scenarios that illustrate how well the datasets represent real-world conditions and the level of difficulty posed by varying environments and obstacles. The results imply that
\begin{itemize}
    \item The two databases include diverse environmental conditions, sufficient to resemble practical use.
    \item Despite the diversity, the models can still learn generalized features that distinguish between different classes.
\end{itemize}
This supports the potential for successful knowledge transfer across different domains in our transfer learning scenarios.

\subsection{Transfer Learning Scenarios}

The source domain comprises these broader datasets, while target domain datasets focus on specific environments or reduced label resolutions.
In particular, we implement \textbf{three transfer learning scenarios}:
\begin{enumerate}
    \item \textbf{General to Specific Room Environments}: Broad data trained and fine-tuned on specific rooms.
    \item \textbf{High to Low Label Resolutions}: Data with high-resolution labels trained and fine-tuned on low-resolution labeled data.
    \item \textbf{Mixed to Specific Environment}: Data combining indoor and outdoor environments trained and fine-tuned on specific room settings.
\end{enumerate}

In our experiments, we employ \textbf{two different types of transfer learning strategies} to evaluate the flexibility and adaptability of the IIns-VAE model with MRC classifiers:
\begin{enumerate}
    \item \textbf{Final Layer Tuning (Classifier Layer Integration)}: In this approach, the entire IIns-VAE model is pre-trained on the source domain, but only the final classifier layer (which outputs the environmental label) is fine-tuned on the target domain. This strategy minimizes the computational load and data requirement for fine-tuning, focusing on adjusting the decision boundaries in the target domain. The final layer was either kept neural-based or replaced by trainable MRC classifiers. The latent variable dimensions for the layer integration strategy were set to $16$.
    \item \textbf{Full Classifier Tuning (Bottleneck Integration)}: In this approach, the pre-trained IIns-VAE model is used to extract latent features up to the bottleneck, and then the entire classifier attached to the bottleneck is fine-tuned on the target domain. The classifier, which is originally neural-based, could be fully replaced with a trainable MRC classifier. This method resembles fountain transfer learning settings, where more significant changes are applied to the model during fine-tuning, allowing the classifier to adapt more thoroughly to the target domain. The bottleneck integration strategy, the latent variable dimensions were increased to $256$, allowing for more expressive feature representation in the latter case.
\end{enumerate}

These two strategies enable us to explore the benefits of lightweight fine-tuning (final layer tuning) versus more intensive adaptation (full classifier tuning), and evaluate how each approach performs in various transfer learning scenarios.

\subsection{Implementations}

Each model was trained for 20 epochs using the Adam optimizer with a learning rate of 
0.0002
0.0002, a batch size of $128$, and a dataset split factor of $0.8$ for training and testing. The IIns-VAE architecture consists of $4$ downsample layers and $3$ residual blocks, providing sufficient depth for capturing relevant features in the wireless signal data. The latent variable dimensions for the layer integration strategy were set to $16$, while for the bottleneck integration strategy, the latent variable dimensions were increased to $256$, allowing for more expressive feature representation in the latter case.

The modules, composing both the baseline IIns-VAE and its variants with MRCs, are built using the following architectures:
\begin{itemize}
    \item \textbf{Encoder}: It includes filters and residual blocks for feature extraction from input signals. The encoder architecture is consistent across all experiments, with adjustable parameters such as the number of filters and downsampling layers.
    \item \textbf{Decoder}: It mirrors the encoder structure, reconstructing the signal from the latent representation.
    \item \textbf{Classifier Base}: Responsible for performing initial environmental classification, directly mapping the latent space of the environment dimension to an initial feature set. Only used for the layer integration cases.
    \item \textbf{Layer Classifier}: Handles classification tasks by further refining the output from the latent space or performing a direct classification when fine-tuning only the final layer. Can be either neural or MRCs.
\end{itemize}


According to the evaluation criteria in \ref{sec:problem_evaluation}, we present the comparative analysis of the baseline methods against our proposed models. Our models integrate MRC with IIns-VAE using both layer-wise and bottleneck-wise transfer learning strategies. This comparison allows us to quantify the benefits of transfer learning and to illustrate how much it improves the performance of environmental identification in dynamic and heterogeneous conditions.

\subsection{Evaluation Criteria: Metrics and Baselines}
\label{sec:problem_evaluation}


We use the $4$ standard metrics to evaluate the performance: accuracy, precision, recall, and F1 score. Each of these metrics captures different aspects of model performance, providing a comprehensive view of the classifier's effectiveness.
\begin{itemize}
    \item \textbf{Accuracy} ($\operatorname{A}$): Measures the overall correctness of the model by calculating the ratio of correctly predicted samples to the total number of samples:
    \begin{equation}
        \operatorname{A} = \frac{\operatorname{TP} + \operatorname{TN}}{\operatorname{TP} + \operatorname{TN} + \operatorname{FP} + \operatorname{FN}}
    \end{equation}
    \noindent where $\operatorname{TP}$ and $\operatorname{TN}$ represent the true positives and true negatives, respectively, and $\operatorname{FP}$ and $\operatorname{FN}$ represent the false positives and false negatives.
    \item \textbf{Precision} ($\operatorname{P}$): Focuses on the proportion of correctly identified positive instances (true positives) out of all instances predicted as positive:
    \begin{equation}
        \operatorname{P} = \frac{\operatorname{TP}}{\operatorname{TP} + \operatorname{FP}}
    \end{equation}
    It reflects the model's ability to minimize false positives.
    \item \textbf{Recall} ($\operatorname{R}$): Also known as sensitivity, recall measures the proportion of actual positive instances that are correctly identified by the model:
    \begin{equation}
        \operatorname{R} = \frac{\operatorname{TP}}{\operatorname{TP} + \operatorname{FN}}
    \end{equation}
    It captures the model's ability to minimize false negatives.
    \item \textbf{F1 Score} ($\operatorname{F1}$): The harmonic mean of precision and recall, providing a balance between the two metrics:
    \begin{equation}
        \operatorname{F1} = 2 \times \frac{\operatorname{P} \times \operatorname{R}}{\operatorname{P} + \operatorname{R}}
    \end{equation}
    It is especially useful when dealing with imbalanced datasets, as it considers both precision and recall.
\end{itemize}

Overall, accuracy provides broad view, while precision and recall focus on specific types of errors (false positives and false negatives). F1 score helps balance these trade-offs, particularly in cases where the dataset is not evenly distributed. These metrics complement each other, ensuring the model performs well under various practical conditions.


We further establish a comparison framework between the proposed transfer learning methods and traditional baseline approaches. The goal is to show how transfer learning improves performance, when compared to existing methods that do no incorporate transfer learning.

The baseline methods are:
\begin{enumerate}
    \item \textbf{Physical Features (PFs) + SVM}: This tradiational ML method uses hand-crafted physical features (PFs) extracted from waveform data, such as received signal strength (RSS), time of arrival (TOA), and advanced metrics like maximum amplitude (MA), rise time (RT), and mean excess delay (MED) \cite{WymMarGifWin:J12}. These features are input into a Support Vector Machine (SVM) classifier for environmental identification. 
    \item \textbf{Physical Features (PFs) + MRC (Linear and Fourier)}: These methods improve upon the PFs + SVM approach by replacing the SVM classifier with a MRC. The MRC, using linear or Fourier-based feature mappings, adds robustness to environmental uncertainties. 
    \item \textbf{IIns-VAE Without Transfer Learning}: This baseline tests the DL-based IIns-VAE model without applying any transfer learning techniques. The model is trained solely on the source domain and directly tested on the target domain.
    It serves as a crucial benchmark, showing the impact of transfer learning when applied to environmental identification tasks.
\end{enumerate}
By incorporating these baseline methods into our evaluation, we establish a clear distinction between traditional approaches and our proposed transfer learning framework. The performance of these methods will help emphasize the necessity of transfer learning in real-world applications where environmental conditions vary significantly between source and target domains.

\subsection{Results Comparison: IIns-VAE vs. MRCs across Integration Strategies}

\begin{table*}[htbp]
\centering
\caption{Performance Comparison for General to Specific Room (LOS/NLOS Classification  Task).}
\begin{tabular}{|l|l|c|c|c|c|}
\hline
\textbf{Model Type} & \textbf{Method} & \textbf{Accuracy} & \textbf{Precision} & \textbf{Recall} & \textbf{F1 Score} \\ \hline
Baselines (No Transfer) & PFs + SVM  &0.840  &0.853  &0.840  &0.845  \\
\cline{2-6}  & PFs + MRC  &0.834  &0.875  &0.834  &0.843 \\
\cline{2-6}  & IIns-VAE &0.541 &0.351 &0.351 &0.351  \\ \hline
Layer Integration
& IIns-VAE        & 0.818             & 0.809              & 0.627           & 0.651             \\ \cline{2-6}
& IIns-VAE+  & \textbf{0.875}             & \textbf{0.883}              & \textbf{0.875}           & \textbf{0.864}             \\ \hline
Bottleneck Integration
& IIns-VAE        & 0.829             & 0.790              & 0.676           & 0.705             \\ \cline{2-6}
& IIns-VAE+  & 0.827             & 0.827              & 0.827           & 0.826             \\ \hline
\end{tabular}
\label{tab:scenario1}
\end{table*}

\begin{table*}[htbp]
\centering
\caption{Performance Comparison for High to Low Label Resolutions ($5$ Material Labels Classification Task).}
\begin{tabular}{|l|l|c|c|c|c|}
\hline
\textbf{Model Type} & \textbf{Method} & \textbf{Accuracy} & \textbf{Precision} & \textbf{Recall} & \textbf{F1 Score} \\ \hline
Baselines (No Transfer) & PFs + SVM  &0.240 &0.499 &0.240 &0.295  \\
\cline{2-6}  & PFs + MRC  &0.267 &0.394 &0.267 &0.249  \\
\cline{2-6}  & IIns-VAE &0.135 &0.027 &0.200 &0.048  \\
\hline
Layer Integration
& IIns-VAE        & 0.420             & 0.199              & 0.276           & 0.223             \\ \cline{2-6}
& IIns-VAE+  & \textbf{0.524}             & \textbf{0.465}              & \textbf{0.524}           & \textbf{0.486}             \\ \hline
Bottleneck Integration
& IIns-VAE        & 0.387             & 0.189              & 0.219           & 0.143             \\ \cline{2-6}
& IIns-VAE+  & 0.381             & 0.426              & 0.381           & 0.395             \\ \hline
\end{tabular}
\label{tab:scenario2}
\end{table*}

\begin{table*}[htbp]
\centering
\caption{Performance Comparison for Mixed to Specific Environment ($5$ Material Labels Classification Task).}
\begin{tabular}{|l|l|c|c|c|c|}
\hline
\textbf{Method Type} & \textbf{Method} & \textbf{Accuracy} & \textbf{Precision} & \textbf{Recall} & \textbf{F1 Score} \\ \hline
Baselines (No Transfer) & PFs + SVM  &0.489 &0.394 &0.489 &0.410  \\
\cline{2-6}  & PFs + MRC  &0.457 &0.351 &0.457 &0.378  \\ 
\cline{2-6}  & IIns-VAE &0.383 &0.176 &0.210 &0.132  \\
\hline
Layer Integration
& IIns-VAE        & 0.323             & 0.124              & 0.181           & 0.124             \\ \cline{2-6}
& IIns-VAE+  & {0.510}             & \textbf{0.536}              & \textbf{0.500}           & \textbf{0.503}             \\ \hline
Bottleneck Integration
& IIns-VAE        & \textbf{0.526}             & 0.330              & 0.384           & 0.336             \\ \cline{2-6}
& IIns-VAE+  & 0.500             & {0.536}              & {0.500}           & {0.503}             \\ \hline
\end{tabular}
\label{tab:scenario3}
\end{table*}





\begin{figure}[htbp]
    \centering
    \subfloat[General to specific room scenario]{%
        \includegraphics[width=0.45\textwidth]{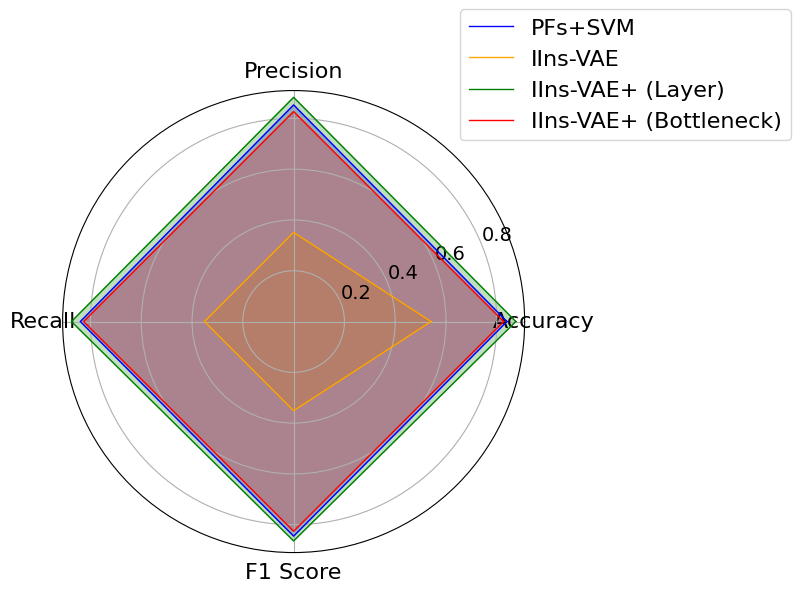}%
        \label{fig:radar_s1}
    }
    \hfill
    \subfloat[High to low label resolution scenario]{%
        \includegraphics[width=0.45\textwidth]{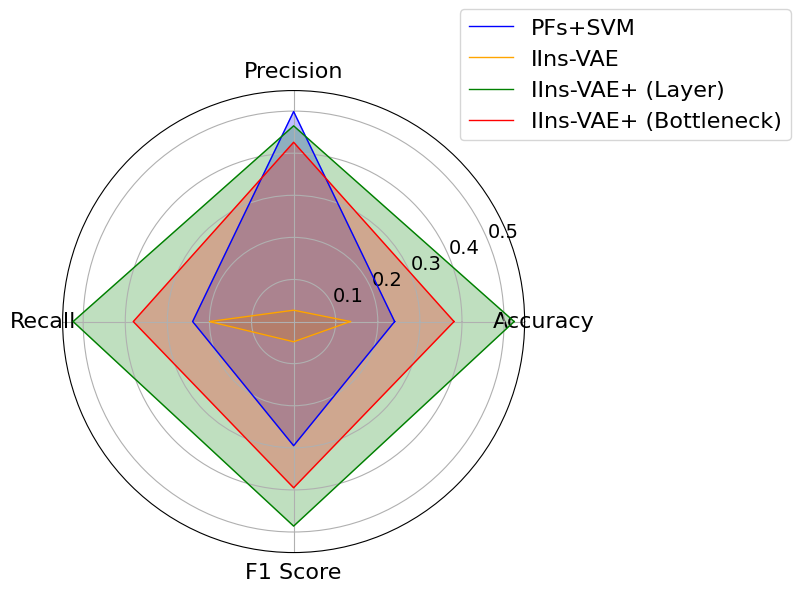}%
        \label{fig:radar_s2}
    }
    \hfill
    \subfloat[Mixed to Specific Environment Scenario]{%
        \includegraphics[width=0.45
\textwidth]{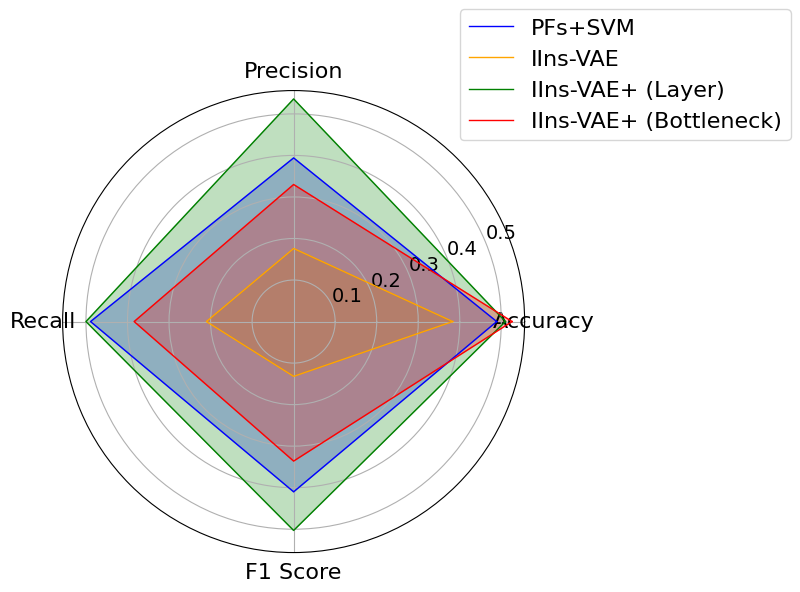}%
        \label{fig:radar_s3}
    }
    \caption{Performance comparison for the three models in terms of the accuracy, precision, recall, and F1 score. The results illustrate the better adaptability of IIns-VAE+ models, especially IIns-VAE+ (Layer), in handling diverse environments with limited target domain data.}
    \label{fig:radar_comparison}
\end{figure}

We present the results of our comparative analysis of IIns-VAE and IIns-MRC models under two distinct integration strategies: layer integration and bottleneck integration, across three different transfer learning scenarios.

\subsubsection{General to Specific Room Environment}

In this scenario, the model is trained on general datasets containing multiple environments and then fine-tuned on a more specific room setting.
Specifically, the source domain is from Database 1-1, containing $9000$ samples recorded in two adjacent office rooms and a parallel connecting hallway. The target domain is a subset from Database 1-2, focusing on a specific office room similar to the one in Database 1-1.
This simulates a common transfer learning challenge where a system is first exposed to diverse environments and subsequently needs to adapt to a narrower context with limited data. The classification task is to distinguish between LOS and NLOS conditions.

As shown in Table \ref{tab:scenario1}, traditional methods using hand-crafted PFs combined with SVM or MRC classifiers perform relatively good in this scenario. IIns-VAE without fine-tuning underperforms compared to bothe tranditional and transfer learning approaches, achieving an accuracy of $0.541$. This implies that IIns-VAE without transfer learning experiences overfitting to the source domain data distribution.

When only the final layer is fine-tuned, we observe that IIns-VAE+ achieves the best overall performance, with an accuracy of $0.875$, precision of $0.883$, recall of $0.875$, and F1 score of $0.864$. IIns-VAE follows with $0.818$ accuracy and $0.651$ of F1 score.

Fine-tuning from the bottleneck level, which allows the entire classifier to be adapted to the target environment, provides a significant performance boost for all methods. IIns-VAE+ still delivers strong results with an accuracy of $0.827$ and an F1 score of $0.826$. Interestingly, IIns-VAE also shows an improvement in accuracy ($0.829$) and F1 score ($0.705$), outperforming its performance in layer integration.

The results clearly show that transfer learning dramatically improves performance compared to no transfer baselines. Both layer and bottleneck integration strategies provide significant gains over the baselines, with bottleneck integration generally outperforming layer integration in terms. This suggests that richer feature representations are critical for the LOS/NLOS classification task in this scenario.

\subsubsection{High to Low Label Resolution}

In this scenario, we evaluate the transfer learning capability when shifting from a high-resolution labeled dataset ($11$ material labels) to a low-resolution one ($5$ material labels).
Specifically, the source domain is from Database 2, containing $44,233$ samples with signal measurements and $11$ different obstacle labels, i.e., $10$ specific objects blocking the LOS path and $1$ LOS label. The target domain is a subset from Database 2 excluded from the source domain data, containing $500$ samples with reduced label resolution, i.e. the obstacle labels are combined into $5$ different material types as metal, glass, wood, plastic, and the LOS label.
The key task involves classifying materials based on the wireless signals, where and challenge lies in handling the reduced label granularity during the adaptation to the target domain.

The results are shown in Table \ref{tab:scenario2}. The performance of traditional machine learning models using hand-crafted PFs with SVM and MRC classifiers, as well as the IIns-VAE without transfer learning, are relatively poor compared to the proposed models. These baseline methods struggle to adapt without transfer learning, with SVM achieving an accuracy of only $0.240$, and IIns-VAE performing particularly poorly with $0.135$ accuracy and an F1 score of $0.048$. These results highlight the limitations of traditional models and non-transfer learning methods when handling the reduction in label resolution.

When only the final layer is tuned, IIns-VAE+ demonstrates the best performance with an accuracy of $0.524$, a precision of $0.465$, recall of $0.524$, and and F1 score of $0.489$. IIns-VAE, on the other hand, struggles in this setting, achieving the lowest performance metrics across the board, with an accuracy of $0.420$ and an F1 score of $0.223$. This suggests that the IIns-VAE architecture is less adept at handling the challenge posed by the reduced label resolution without deeper integration of the domain information.

Similar to the previous scenario, fine-tuning from the bottleneck level slightly improves performance for both methods, although the gains are less pronounced in this case. Interestingly, IIns-VAE+ does not benefit as much from bottleneck integration in this scenario, achieving slightly lower scores than with layer integration.

Transfer learning shows clear improvements over the non-transfer baselines. Bottleneck integration continues to improve performance, but the complexity of the low-label resolution scenario reduces the advantage compared to the first scenario. IIns-VAE+ remains the most robust across both integration types. This scenario further illustrates the robustness of IIns-VAE+ model and highlights the need for careful integration strategy selection based on task complexity.

\subsubsection{Mixed to Specific Environment}

In this scenario, we explore the transfer learning challenge of moving from a mixed environment to a more specific setting, where the goal is to classify materials in distinct environments.
Specifically, the source domain includes data from Database 2, containing $44,233$ samples measured in the big room, small room, and outdoor scenario. The target domain is a subset of data excluded from Database 2, containing $5000$ samples measured from the medium office room scenario.
The task emphasizes how well the models can adapt to narrower, more focused contexts after being trained on broader environmental data. The results are presented in Table \ref{tab:scenario3}.

The baseline methods (PFs + SVM and PFs + MRC) demonstrate moderate performance, with SVM achieving an accuracy of $0.489$ and F1 score of $0.410$. IIns-VAE trained only on the source domain without transfer yields a lower accuracy of $0.383$ and F1 score of $0.132$, showing the clear need for transfer learning in this scenario.

When only the final layer is tuned, IIns-VAE+ shows the best performance, achieving an accuracy of $0.500$, a precision of $0.536$, and an F1 core of $0.503$. On the other hand, IIns-VAE performs poorly in this scenario, with an accuracy of only $0.323$ and an F1 score of $0.124$, reflecting the challenges of adapting its architecture to the specific environment.

Fine-tuning from the bottleneck level continues to show improvements for IIns-VAE, which achieves the highest accuracy ($0.526$) and an F1 score of $0.336$. However, IIns-VAE+ performs similarly across both integration strategies, with Fourier integration again slightly outperforming linear in terms of precision and F1 score ($0.536$ and $0.503$, respectively). Notably, in this scenario, the bottleneck integration for IIns-VAE yields better performance compared to layer integration. This suggests that deeper adaptation is necessary when moving from mixed to specific environments.

In this scenario, IIns-VAE+ again demonstrate consistent performance across both integration methods, while IIns-VAE benefits more from bottleneck integration. The need for deeper adaptation when moving from mixed environments to specific ones is evident, as bottleneck integration provides more flexibility in adjusting to the target domain. Overall, the results highlight the importance of the MRC's advanced classification mechanisms in scenarios that demand high adaptability.

\subsubsection{Cross-Scenario Comparison and Discussion}

The performance comparison across transfer learning scenarios is summarized in Figures \ref{fig:radar_comparison}. Several key insights emerge:
\begin{enumerate}
    \item \textbf{General to Specific Room Scenario}: As shown in Figure \ref{fig:radar_s1}, this scenario achieves the strongest overall performance across methods. IIns-VAE+ (Layer integration) outperforms all models, with an accuracy of $0.875$ and an F1 score of $0.864$. These results indicate that the IIns-VAE+ approach can effectively generalize to specific room environments when pre-trained on general data, likely due to its ability to capture complex spatial patterns. Layer integration appears particularly suitable here, as it retains finer feature details without extensive adaptation, which is beneficial in structured, room-based environments.
    \item \textbf{High to Low Label Resolution}: Figure \ref{fig:radar_s2} shows that performance drops for all models in this challenging scenario due to the reduced label resolution. Despite this, IIns-VAE+ (Layer integration) maintains a competitive edge, with an accuracy of $0.524$ and an F1 score of $0.486$, outperforming other models in distinguishing between classes. This highlights the strength of Fourier-based feature mappings in preserving meaningful class distinctions even under low-resolution labels, though the performance gap between layer and bottleneck integration narrows, given the inherent limitations of the low-resolution environment.
    \item \textbf{Mixed to Specific Environment}: In the more variable environment of Figure \ref{fig:radar_s3}, IIns-VAE+ (Bottleneck integration) achieves the highest accuracy ($0.526$) and shows a significant improvement in precision, recall, and F1 score over other methods. Bottleneck integration's deeper adaptation enables IIns-VAE+ to capture complex global representations, enhancing its robustness against variability in this mixed-to-specific setting. This adaptability proves essential in scenarios with high environmental variability, where bottleneck integration allows the model to manage complex domain shifts more effectively.
\end{enumerate}

Across all scenarios, IIns-VAE+ consistently outperforms the traditional models (e.g., PFs+SVM and IIns-VAE alone), particularly in terms of F1 score, precision, and recall. The radar plots in Figure \ref{fig:radar_comparison} illustrate that IIns-VAE+ with bottleneck integration significantly boosts these metrics, especially in unstructured or chaotic environments. This can be attributed to the bottleneck’s role in enabling global feature adaptation, which mitigates the impact of domain shifts.

\begin{figure*}[htbp]
\centering
\includegraphics[width=0.9\textwidth]{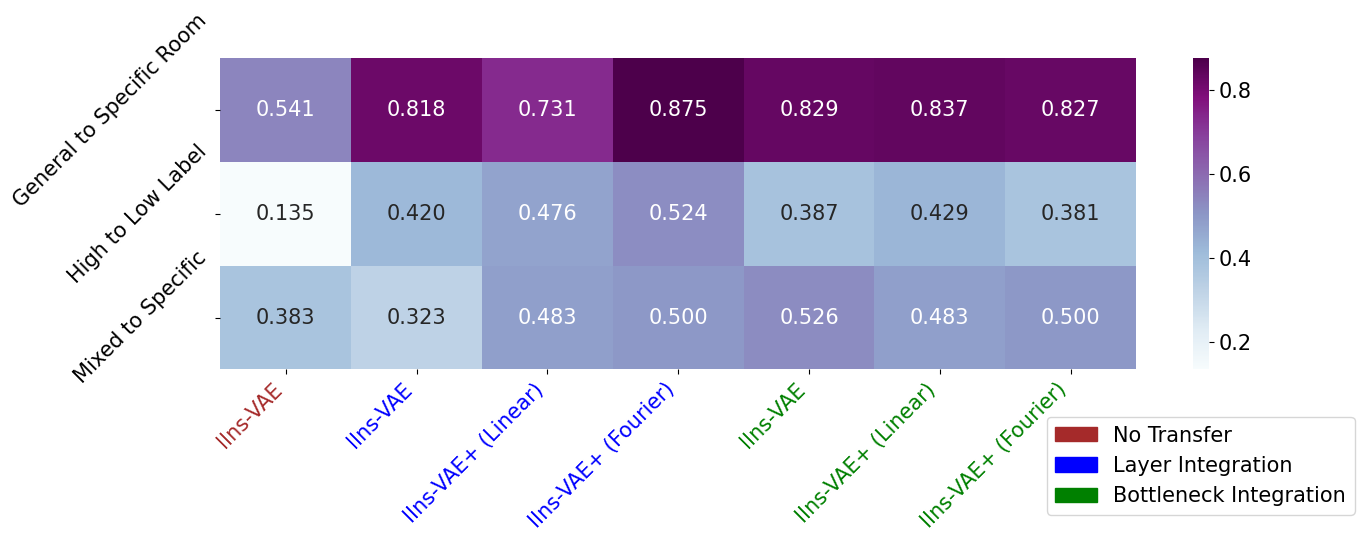}
\caption{Comparison of Accuracy for IIns-VAE with MRCs under layer and bottleneck integration setting across different transfer learning scenarios.}
\label{fig:heatmap_acc}
\end{figure*}

\begin{figure*}[htbp]
\centering
\includegraphics[width=0.9\textwidth]{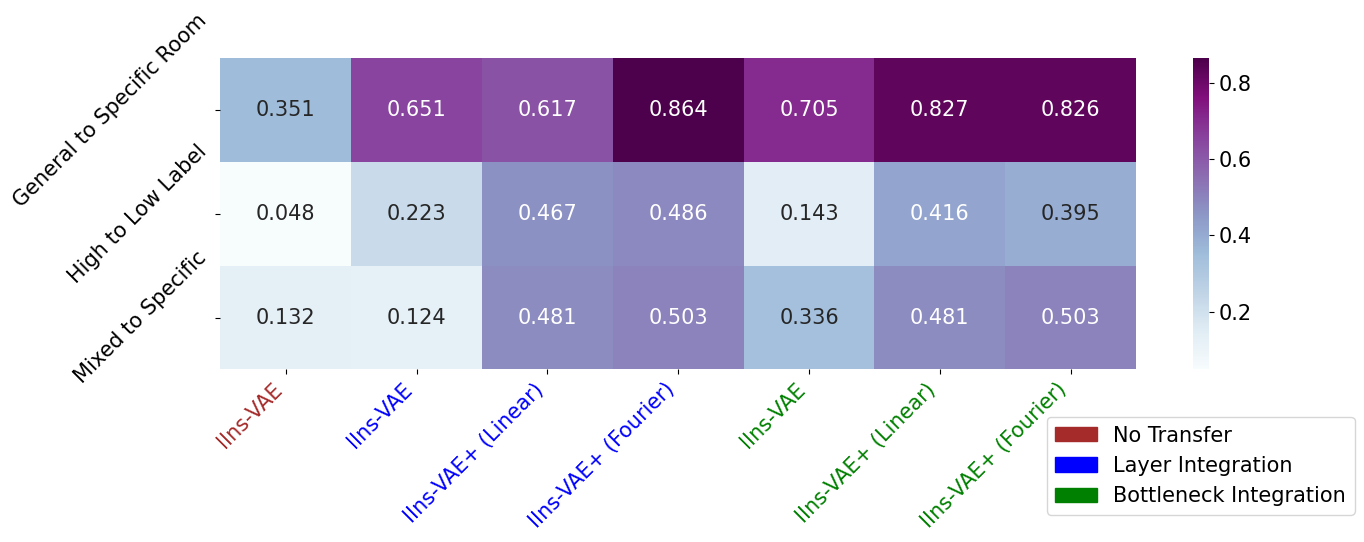}
\caption{Comparison of F1 Score for IIns-VAE with MRCs under layer and bottleneck integration setting across different transfer learning scenarios.}
\label{fig:heatmap_f1}
\end{figure*}

The results also indicate the importance of selecting integration strategies according to the environmental structure. Specifically, layer integration performs best in structured environments (e.g., the general to specific room scenarios), where retaining precise feature details through Fourier mappings supports more accurate classification without extensive adaptation. Bottleneck integration excels in variable or chaotic environments (e.g., the mixed-to-specific scenario), where adapatability in the latent space enhances the model's resilience against domain shifts, improving precision, recall, and F1 score.

In conclusion, the findings demonstrate the value of task-specific integration strategies. For structured environments, IIns-VAE+ with layer integration is highly effective, leveraging fine-grained latent features. Conversely, in more variable environments, IIns-VAE+ with bottleneck integration offers enhanced robustness, benefiting from deep adaptation in the latent space. These results highlight the strengths of combing DL's feature extraction with MRC's risk-aware classification for robust transfer learning across diverse wireless sening tasks.

\subsection{Ablation Study across Different Feature Mappings}

In addition to the primary results leveraging F feature mappings for the MRC classifiers, we conduceted an ablation study to evaluate the performance impact of using linear feature mappings. This comparison provides insight into the flexibility and adaptability of MRC when paired with simpler or more complex mappings. The results are shown in heatmaps in \ref{fig:heatmap_acc}-\ref{fig:heatmap_f1}.

The results indicate that Fourier feature mappings consistently outperform linear mappings across all scenarios and metrics. For instance, in the "General to Specific Room" scenario, the accuracy of IIns-VAE+ with Fourier mappings reaches $0.875$ under layer integration, compared to $0.731$ with linear mappings. Similarly, the F1 score for the "High to Low Label Resolution" scenario improves from $0.467$ (linear) to $0.486$ (Fourier) with layer integration.
However, linear mappings still offer competitive performance in less complex settings, such as the "Mixed to Specific Environment" scenario, where they achieve comparable results to Fourier mappings under bottleneck integration.

In summary, the ablation study illustrates the importance of selecting feature mappings based on the complexity of the transfer learning task. Fourier mappings are better suited for scenarios requiring higher adaptability, while linear mappings provide an efficient alternative for simpler environments.



\section{Conclusion}
\label{sec:con}

This paper presents IIns-VAE+, a hybrid framework for environmental identification in wireless sensing that integrates the DL capabilities of IIns-VAE with the robust classification of MRC. Experimental results demonstrate that IIns-VAE+ significantly enhances generalization across diverse environments, outperforming both traditional methods and the baseline IIns-VAE model in various transfer learning scenarios.

Our main contribution is a two-stage framework designed to address the challenge of environmental identification across domains. The framework first trains IIns-VAE in the source domain to extract latent environmental features, and then trains MRC in the target domain to adapt the classifier to unseen environments. This decoupled approach provides a practical solution for enhancing environmental perception in ISAC systems. It directly enables the context-aware intelligence required for dynamic optimization of resource allocation, beamforming, and path planning in real-world, non-stationary wireless environments.

Consequently, our work offers a robust solution to the problem of domain shift in practical wireless sensing for 6G networks. Future work will investigate more advanced machine learning, particularly deep learning techniques, to further strengthen the proposed framework and extend its applicability to a broader spectrum of ISAC tasks.

\ifCLASSOPTIONcaptionsoff
  \newpage
\fi

\end{document}